\documentclass{article} 
\usepackage{iclr2024_conference,times}
\usepackage{physics}

\usepackage{amsmath,amsfonts,bm}

\def\eqref#1{equation~\ref{#1}}

\def\1{\bm{1}}

\def\ve{{\bm{e}}}

\DeclareMathAlphabet{\mathsfit}{\encodingdefault}{\sfdefault}{m}{sl}
\SetMathAlphabet{\mathsfit}{bold}{\encodingdefault}{\sfdefault}{bx}{n}

\newcommand{\Model}[1]{SpaTea}

\usepackage{hyperref}
\usepackage{url}
\usepackage{graphicx}
\usepackage{float}

\usepackage{adjustbox}

\usepackage{booktabs}       
\usepackage{amsfonts,amsmath,amssymb,amsthm}
\usepackage{multicol, multirow}

\title{A Quantum-Inspired Neural Network for Geometric Modeling}
\author{%
\small
Weitao Du$^*$\\
Chinese Academy of Sciences\\
Huawei Technologies Ltd\\
\textit{duweitao@amss.ac.cn}\\
\And
Shengchao Liu$^*$\\
Université de Montréal\\
\textit{shengchao.liu@umontreal.ca}\\
\And
Xuecang Zhang$^*$\\
Huawei Technologies Ltd\\
\textit{zhangxuecang@huawei.com}
}

\iclrfinalcopy 
\begin{document}

\maketitle

\begin{abstract}
By conceiving physical systems as 3D many-body point clouds, geometric graph neural networks (GNNs), such as SE(3)/E(3) equivalent GNNs, have showcased promising performance. In particular, their effective message-passing mechanics make them adept at modeling molecules and crystalline materials. However, current geometric GNNs only offer a mean-field approximation of the many-body system,  encapsulated within two-body  message passing, thus falling short in capturing intricate relationships within these geometric graphs. To address this limitation, tensor networks, widely employed by computational physics to handle many-body systems using high-order tensors, have been introduced. Nevertheless, integrating these tensorized networks into the message-passing framework of  GNNs faces  scalability and symmetry conservation (e.g., permutation and rotation) challenges. In response, we introduce an innovative equivariant Matrix Product State (MPS)-based message-passing strategy, through achieving an efficient implementation of the tensor contraction operation. Our method   effectively models complex many-body relationships, suppressing mean-field approximations, and captures symmetries within geometric graphs. Importantly, it seamlessly replaces the standard message-passing and layer-aggregation modules intrinsic to geometric GNNs. We empirically validate the superior accuracy of our approach on benchmark tasks, including predicting classical Newton systems and quantum tensor Hamiltonian matrices. 
To our knowledge, our approach represents the inaugural utilization of parameterized geometric tensor networks.

\end{abstract}

\section{Introduction}

The conceptualization of physical systems as dynamic 3D many-body point clouds has instigated a transformative shift in learning to model graph structured data. In this innovative paradigm, geometric neural networks (GNNs)~\citep{schutt2018schnet,thomas2018tensor}, notably exemplified by SE(3)/E(3) equivalent GNNs~\citep{schutt2021equivariant,liao2022equiformer,satorras2021n,du2022se},  have emerged as powerful tools, demonstrating exceptional performance through effectively capturing atomic interactions in molecules and crystalline materials. Notably, their efficacy is prominently attributed to the deployment of effective message-passing mechanics, rendering them well-suited for modeling the intricate structures in these geometric data~\citep{liu2023symmetry}. 

Despite their promising performance, contemporary geometric GNNs are undesirably encapsulated within two-body message passing mechanism~\citep{10.5555/3305381.3305512}. That is, the messaging paradigm in these geometric GNNs, despite its computational expediency, confines the expressive capabilities of the GNN models due to its reliance on a mean-field approximation of many-body interaction~\citep{10.1093/acprof:oso/9780198743736.001.0001,barbier2015statistical}. This limitation becomes particularly evident when modeling intricate relationships involving more than two bodies, as is the case for atomic interactions in molecules and materials. Thus,  empowering current geometric GNNs to achieve comprehensive expressiveness necessitates a paradigm shift towards effective modeling of the high-order interactions of such systems. 

To attain the aforementioned goal, we leverage tensor networks, a well-established tool in computational physics for representing intricate many-body interactions using high-order tensors~\citep{kottmann2022investigating}, which are multi-dimensional arrays of numbers~\footnote{Intuitively, tensors generalize scalars (order-0 tensors), vectors (order-1 tensors), and matrices (order-2 tensors) to higher dimensions or orders. A tensor network is constructed using a set of tensors, where certain tensor indices are interconnected to form the network structure.}. 
These tensor networks are developed to effectively represent quantum states and capture transformation between these  states in a physical system. Furthermore, the ability to compress high-order tensors by truncating the virtual dimension between tensors makes tensor networks an attractive solution for modeling large systems. Nevertheless, seamlessly integrating tensorized networks into the message-passing framework of geometric GNNs  encounters issues of scalability and preservation of permutation and rotation symmetry. To address these challenges, we devise a novel approach called Spatial and Temporal Tensor Network Aggregation (SpaTea). In specific, it leverages a novel equivariant Matrix Product State (MPS)-based message-passing strategy (1-dimensional tensor network ~\citep{10.5555/2011832.2011833, articleVerstraete}), capitalizing on a scalable and efficient implementation of the tensor contraction operation~\footnote{Tensor contraction generalizes both matrix multiplication and vector inner product. It involves multiplying two tensors along a pair of indices that share equal dimensions.}  embedded within the deep GNN framework. This contraction effectively reduces the tensors' dimensions, enhancing computational efficiency.  
Significantly, drawing inspiration from the density matrix renormalization group, our method offers a unified perspective on traditional message-passing neural networks (MPNNs) concerning nodes and layers. As a result, our strategy seamlessly replaces the standard message-passing and layer-aggregation modules intrinsic to geometric GNNs. This seamless integration of tensorized networks elevates the geometric model's expressive capacity without disrupting the conventional geometric deep GNN architecture. Different from sub-graph based graph neural networks which aggregates information on the subgraph level, our basic unit is still in the node level. On the other hand, from the data point view, the existence many-body entanglement between tokens inside the data has been demonstrated, e,g, CV data~\citep{10.3389/fams.2021.716044}, NLP data~\citep{pestun2017tensor}, materials~\citep{sommer2022entangling}, which makes the tensor-network based modeling natural. 

 We empirically validate the performance of our method \Model{} on benchmark tasks, including predicting classical Newton systems and quantum tensor Hamiltonian matrices. 
 Our results establish new state-of-the-art accuracy for these specific tasks.
To our knowledge, \Model{} represents the first   utilization of  parameterized geometric tensor networks. 
Also, it is worth noting that, owing to its tensor network foundation, \Model{} holds the potential to evolve into genuine quantum learning algorithms tailored for geometric graphs, offering potential advantages for quantum-related tasks. 

We summarize our main contributions as follows. 
\begin{itemize}
    \item Identify and resolve the mean-field message passing limitation in  geometric GNNs. 
    \item Devise the first  parameterized geometric tensor networks for geometric graph modeling. 
    \item Empirically  show our  method's superior accuracy on benchmark   tasks.  
\end{itemize}

\section{Preliminaries}
\begin{figure}[tb!]
\centering
\includegraphics[width=0.8\textwidth]{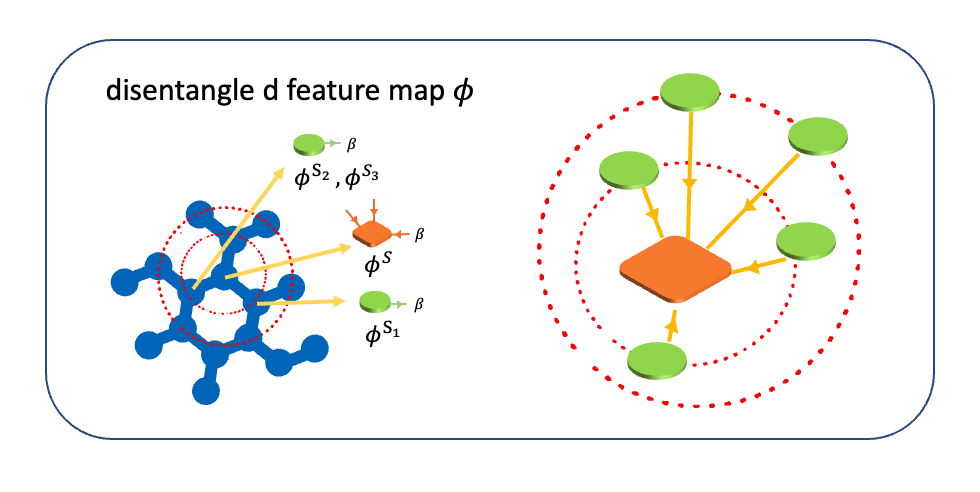}
\vspace{-4ex}
\caption{The initial step of spatial mixing: organizing neighborhood embeddings according to their (Euclidean) distance, e.g., $\phi^{s_3} \rightarrow \phi^{s_2}$ and $\phi^{s_1} \rightarrow \phi^{s}$. When $\phi^{s_2}$ and $\phi^{s_1}$ are equidistant from $\phi^s$, the spatial mixing process should be permutation-invariant when the order of $s_1$ and $s_2$ is exchanged.}
\label{fig:spatial_mixing1}
\vspace{-1ex}
\end{figure}
\subsection{Key Components of Message Passing Neural Networks}
The foundation of message-passing neural networks can be traced back to a model that represents structured data as a sequence of tokens. This concept has evolved into a unified framework where data processing can be viewed as an attention mechanism between tokens, especially from a transformer's perspective. For instance, in the context of a 2D image, tokens are defined as local patches within the image, while in natural language processing (NLP), tokens are individual words or characters. Both scenarios can be understood as ordered graphs with fully connected edges linking the tokens. However, it is noteworthy that in this paper, we specifically focus on an entirely different paradigm: unordered graphs, denoted as $(V, E)$, where $V$ denotes the vertices and $E$ represents the edge features. These unordered graphs are instrumental in the context of our tensor network algorithm, which draws an analogy to a many-body interaction system as we develop our method. 

We now introduce the ``classical" message passing scheme \citep{gilmer2017neural} from layer $t-1$ to $t$, which is also equivalent to the token mixing module introduced in \citep{guibas2021adaptive} when the graph is a regular mesh. This scheme also encompasses the attention mechanism as a special case:
\begin{equation} \label{eq: mpnn}
x_i^t = \sum_{x_j \in N(x_i^{t-1})} W_{ij}(x_i^{t-1}, x_j^{t-1}) \phi(x_i^{t-1}, x_j^{t-1}),
\end{equation}
where the neighbors of $N(x_i)$ are determined by the adjacency matrix, 
and $W_{ij}$, $\phi$ are matrix-, vector-valued functions parameterized by multi-layer neural networks (MLP).   It's worth noting that the summation $\sum$ can be replaced with other permutation pooling operations, such as maximum or minimum. By summarizing $W_{ij}$ and $\phi$, we can broadly state that the representation of each node, $x_i^t$, is updated using an operator $H_{N(x_i^{t-1})}$ that depends on the local neighborhood of each node $x_i$ for each layer:
\begin{equation} \label{eq: mpnn operator}
x_i^t = H_{N(x_i^{t-1})} (x_i^{t-1}).
\end{equation}
However, from a statistical physics perspective, Equation \ref{eq: mpnn} serves as a mean-field approximation of the many-body interaction system. In other words, only information that is in the range of mean-field approximation can be obtained in the form of \ref{eq: mpnn}. As this mixing occurs at the token level, we refer to this process as \textbf{spatial aggregation}.  Note that in standard GNN pipeline, there is also an additional node update layer for transforming $x_i^t$. For example, in a standard graph transformer architecture (e.g., \cite{kreuzer2021rethinking}), node update is formulated as
\begin{equation}x_i^t = \text{Layer normalization}(x_i^t) + W x_i^{t-1}. \end{equation} 
In addition to spatial aggregation, \cite{xu2018representation} observed that the output representations of each layer contain varying hierarchical information. For example, shallower layers contain more local information. Therefore, \cite{xu2018representation} proposed various methods for aggregating the outputs of each layer, rather than simply pooling the representation of the final layer as the output (see also \cite{xupowerful,guo2021sequential}). Since we index the layers with the variable $t$, we refer to this aggregation as \textbf{temporal aggregation}.

Temporal aggregation involves a dynamic process of merging information from different layers. One common approach is to apply a function, denoted as $\mathcal{A}_t$, that aggregates the representations from all layers up to time step $t$. This can be formulated as:
\begin{equation} \label{eq: layer agg}
x_i^t = \mathcal{A}_t\left(x_i^{0}, x_i^{1}, \ldots, x_i^{t-1}\right).
\end{equation}
The choice of the aggregation function $\mathcal{A}_t$ can vary and may depend on the specific problem at hand. It is designed to capture the different information across different layers, with the flexibility to emphasize contributions from earlier or later stages of the computation. 


\subsection{Tensor Network as an Efficient Tensor Representation}
To efficiently represent complex many-body systems, we turn to tensor networks, a powerful framework used to mitigate the challenges posed by high-order tensors appeared both as a representation of quantum states and the transformation between quantum states \citep{ran2020tensor}. To facilitate our discussion, we adopt the Dirac ket and bra notation to represent vectors and their dual vectors. Consider a basis $\{ | s \rangle \}$ for a $d$-dimensional vector space $H$, which represents the single-body state space. The N-body (entangled) state space is the repeated tensor product of $H = H_0 \otimes H_1 \otimes \cdots \otimes H_{N-1}$, where each $H_i$ is an identical copy of the high dimensional vector space $H$. In other words, we only consider many-body system consisted of bosonic particles. In this space, the basis is given by:
\begin{equation} \label{eq: polynomial state}
| s_0 ,s_1,\dots, s_{N-1} \rangle\
: = \{ | s_0 \rangle \otimes \cdots \otimes | s_{N-1} \rangle\},\ \text{where } s_0, \ldots, s_{N-1} = 0, 1, \ldots, d-1.
\end{equation}
Any N-body state can be expressed as $| \phi \rangle = \sum_{s_0, \ldots, s_{N-1}} \phi_{s_0, \ldots, s_{N-1}} | s_0 ,\dots, s_{N-1} \rangle$. The coefficients $\phi_{s_0, \ldots, s_{N-1}}$ constitute a tensor with N indices, leading to the issue of the ``exponential wall" as N grows. Specifically, the dimension of the high-order tensor $\phi_{s_0, \ldots, s_{N-1}}$ becomes $d^N$. This phenomenon is often referred to as the 'curse of dimensionality,' wherein the computational cost and storage requirements become overwhelmingly large as N increases. Once defined the state space as a high rank tensor space, the transformation of quantum states $\mathcal{O}$  is a linear operator, which can also be written in the ket and bra notation:
\begin{equation} \label{eq: linear operator}
\mathcal{O} = \sum_{s_0,\dots,s_{N-1}} O^{s_0s_0^,}\dots O^{s_{N-1}s_{N-1}^,} | s_0 ,\dots, s_{N-1} \rangle \langle s_0 ,\dots, s_{N-1} |.
\end{equation} 
We expect a similar "exponential wall" issue when dealing with an expressive aggregation operator $H_{N(x_i^{t-1})}$ in Equation~\ref{eq: mpnn operator}, as it also depends on the many-body state of the neighbors $N(x_i^{t-1})$. To efficiently tackle this problem, physicists have developed various tensor network state representations for compressing high-order tensors. In our context, we employ a specialized tree tensor network state known as the Matrix Product State (MPS):
\begin{equation}
\phi_{s_0, \ldots, s_{N-1}} = \sum_{a_0, a_1, \ldots, a_{N-2}} A^0_{s_0, a_0} A^1_{a_0, s_1, a_1} \cdots A^{N-1}_{a_{N-2}, s_{N-1}}.
\end{equation}
We call the index $a_i$ as virtual indexes, as they are contracted and absent in the final $\phi_{s_0, \ldots, s_{N-1}}$. On the contrary, $s_i$ is the physics index. We choose the matrix product state for two main reasons: Firstly, it requires less specialized domain knowledge compared to other tensor network states (e.g., Projected Entangled Pair States (PEPS) \citealp{cirac2021matrix}). Secondly, it shares a low-rank approximation property guaranteed by the tensor-train decomposition. While the physical index remains the set $\{s_0, \ldots, s_{N-1}\}$, the dimension $\chi$ of the virtual indices $\{a_0, \ldots, a_{N-2}\}$ can be set as a hyperparameter. As a result, the dimension of an N-order matrix product state becomes:
\begin{equation}
\mathcal{O}(N\chi^2 d). 
\end{equation}
This representation offers an efficient compression of the original high-order tensor when $N > 1$ and $\chi \ll d$. Moreover, it's been mathematically proven that every N-th order tensor has an optimal low-rank approximation in the MPS form \citep{oseledets2011tensor}. 

\begin{table}
\centering
\caption{A non-formal analogy between machine learning and many-body quantum mechanics.}
\begin{tabular}{||c||c||}  
\hline
Quantum Mechanics & Machine Learning \\ [0.5ex] 
\hline\hline
Quantum State & Raw Data \\ 
\hline
Coefficients of Quantum State & Embedding \\
\hline
Hamiltonian Operator & Layer Update \\
\hline
Renormalization & Aggregation and Pooling \\
\hline
Tensor-Product State & Product Probability \\
\hline
Entanglement State & Non-Independent Probability \\ [1ex] 
\hline
\end{tabular}
\label{table:1}
\end{table}

\subsection{Quantum-Inspired Aspects}

In this subsection, we shed light on the quantum-inspired elements that underpin our tensor network-based approach. While our primary focus is on adapting tensor networks for many-body systems within the domain of machine learning, we draw inspiration from quantum principles to address the challenges of high-dimensional data representation.

Quantum physics has long served as a source of inspiration for various computational and machine learning techniques. In our case, the inspiration from quantum mechanics lies in the efficient representation of multi-dimensional data, similar to the way quantum states and operators are expressed in the quantum realm. By harnessing quantum-inspired concepts, we aim to bridge the gap between classical machine learning methodologies for geometric graphs and the complexities of many-body systems.

The integration of Matrix Product State (MPS) into our tensor network architecture, for instance, reflects the quantum-inspired notion of efficiently encoding complex states. As we explore the applications of MPS and other tensor network forms within the context of message-passing neural networks, we aspire to achieve a balance between the elegance of quantum principles and the computational needs of our many-body system models.

Throughout this paper, we will elucidate the specific quantum-inspired aspects of our method and demonstrate how they enhance our ability to model and process high-dimensional data efficiently, offering a unique perspective on the fusion of quantum concepts and machine learning principles within a tensor network framework.



\section{Method}
As we have briefly mentioned in the Introduction section, the universal expressiveness of 3D equivariant Graph Neural Networks (GNNs) hinges on the network's ability to represent and compute equivariant polynomials of arbitrary orders, a fundamental requirement for capturing the complex relationships within geometric graphs. 

To build a neural layer capable of expressing these equivariant polynomials, one might turn to tensorized graph neural networks \citep{maron2018invariant}, which provide a generalization of polynomial neural networks \citep{kileel2019expressive} while preserving permutation invariance. However, we immediately encounter two significant practical challenges:

1. The dimension of the tensor representation, required for learning transformations between these polynomials, scales as $\mathcal{O}(d^N \times d^N)$. In practice, this dimension becomes prohibitively large, necessitating effective truncation techniques.

2. The architecture of tensorized graph neural networks does not align with the spatial and temporal aggregation scheme introduced in Section 2.1, creating a challenge for the expression of higher-order geometric tensors within Equations~\ref{eq: mpnn} and \ref{eq: layer agg} with a well-controlled truncation parameter.

A natural question we may ask is: why do we need to model and compute higher-order geometric tensors? A key insight is captured in the concept of ``frame transition", one of the essential building blocks for constructing expressive equivariant GNNs. The frame transition (FT) encodes critical information, such as the torsion angle between chemical bonds in molecular structures. For instance, in scenarios involving shared atoms connected by k bonds, the computation of frame transitions necessitates a quadratic complexity of $O(k^2)$. Frame transitions, however, can be efficiently represented as rank-two polynomials by introducing node-wise equivariant frames (see Appendix \ref{ap: artitectures}).

FT offers a concrete example to underscore the importance of higher-order geometric tensors in modeling complex relationships within geometric graphs. Moreover, many real-world problems, such as solving the Schrödinger equation in quantum chemistry, involve Hamiltonian operators with many-body interactions up to the fourth order. In the quantum realm, the Density Matrix Renormalization Group (DMRG) algorithm plays a critical role for approximately solving such systems, as well as offering a theoretical guarantee on the accuracy of truncation in complex Hamiltonian operators.

Therefore, the motivation behind the development of our equivariant Matrix Product State (MPS)-based message-passing method, inspired by the Density Matrix Renormalization Group (DMRG) algorithm, lies in addressing the challenge of effectively and efficiently modeling many-body relationships and symmetries in geometric graphs. To embark on this journey, we establish a crucial link between the aggregation mechanisms described in Equations \ref{eq: mpnn} and \ref{eq: layer agg} with the renormalization procedures employed in DMRG.

\subsection{Matrix Product Operator as a General Renormalization Technology}
In this section, we elaborate on the integration of the matrix product state (MPS) to facilitate the realization of the spatial mixing operator $H_{N(x_i^{t-1})}$ and the temporal mixing module.

From Equation~\ref{eq: mpnn operator}, the operator $H_{N(x_i^{t-1})}$ essentially performs a parameterized tensor contraction operation, possibly with non-linearity, utilizing the concatenation of each node's neighbors: $| x_0 \rangle \otimes \cdots \otimes | x_n \rangle$, where $\{| x_0 \rangle, \ldots, | x_n \rangle \}$ represents the vector embeddings of the neighboring nodes. It's worth noting that tensor contractions within the context of a given tensor network structure remain a subject of active research in the physics community. Here, we briefly mention two standard tensor contraction algorithms: 1. Time-Evolving Block Decimation (TEBD, \cite{Suzuki1976GeneralizedTF}); 2. Density Matrix Renormalization Group (DMRG, \cite{white1993density}). 

While both algorithms can be adapted into a deep neural network framework (with time in TEBD roughly corresponding to the layer index in deep learning), DMRG is better suited for fitting the traditional message-passing scheme. More precisely, in TEBD, each contraction operation will double the dimension virtual indexes. Therefore, to derive a node representation from it, a subsequent compression (pooling) is usually required (see \cite{PhysRevLett.93.207204} for an illustration). Conversely, DMRG defines an \textbf{effective Hamiltonian operator} for each node by renormalizing the global information of other body states, aligning well with the form of $H_{N(x_i^{t-1})}$. We provide an illustration in our framework for DMRG in Appendix \ref{AP: DMRG}. Given the complexity of geometric graph scenarios, where spatial mixing necessitates $H_{N(x_i^{t-1})}$ to be both permutation and $SE(3)$ equivariant, we reserve the detailed implementation for the subsequent section.

As for the layer aggregation step \ref{eq: layer agg}, it inherently possesses a renormalization direction from $t-1$ to $t$, proceeding through the layers until the final layer is reached. To implement this module, we adapt the 1-dimensional aResMPS introduced in \cite{meng2023residual}. Specifically, let $v_n^0$ represent the initial node embedding (commonly a one-hot embedding in our experiments), and $x^l$ denote the node representation for the l-th layer. The layer aggregation  formula $\Psi(v_n^0, x_n^1,\dots,x_n^L)$ is defined iteratively as follows:
\begin{equation}  \label{eq: temporal mps}
v_{na_{l+1}}^{l+1} = v_{na_{l}}^{l} + \sigma\left(\sum_{a_ls_l} v_{na_{l}}^{l}x^l_{ns_l}\frac{\phi^l_{a_ls_la_{l+1}}}{Z(l)} + b_{a_{l+1}}^l\right).
\end{equation}
In this equation, the three-order tensor $\phi^l_{a_ls_la_{l+1}}$ represents the parameterized matrix product operator for aggregating the previous representation and the newly incoming representation $x^l_{ns_l}$, while $b_{a_{l+1}}^l$ denotes the bias for the l-th layer. From now on, $\Psi(v_n^0, x_n^1,\dots,x_n^L)$ is named as the \textbf{temporal aggregation kernel}.
Notably, $Z(l): = \lVert x^l_{ns_l}\phi^l_{a_ls_la_{l+1}} \rVert$ is utilized as a normalizing factor to maintain the scale of the aggregation. While it's possible to address the scale exploration problem by multiplying a small $\epsilon$ to $x^l_{ns_l}$, as discussed in \citep{meng2023residual}, our empirical findings favor the presented normalization strategy.

\begin{figure}[tb!]
\centering
\includegraphics[width=\textwidth]{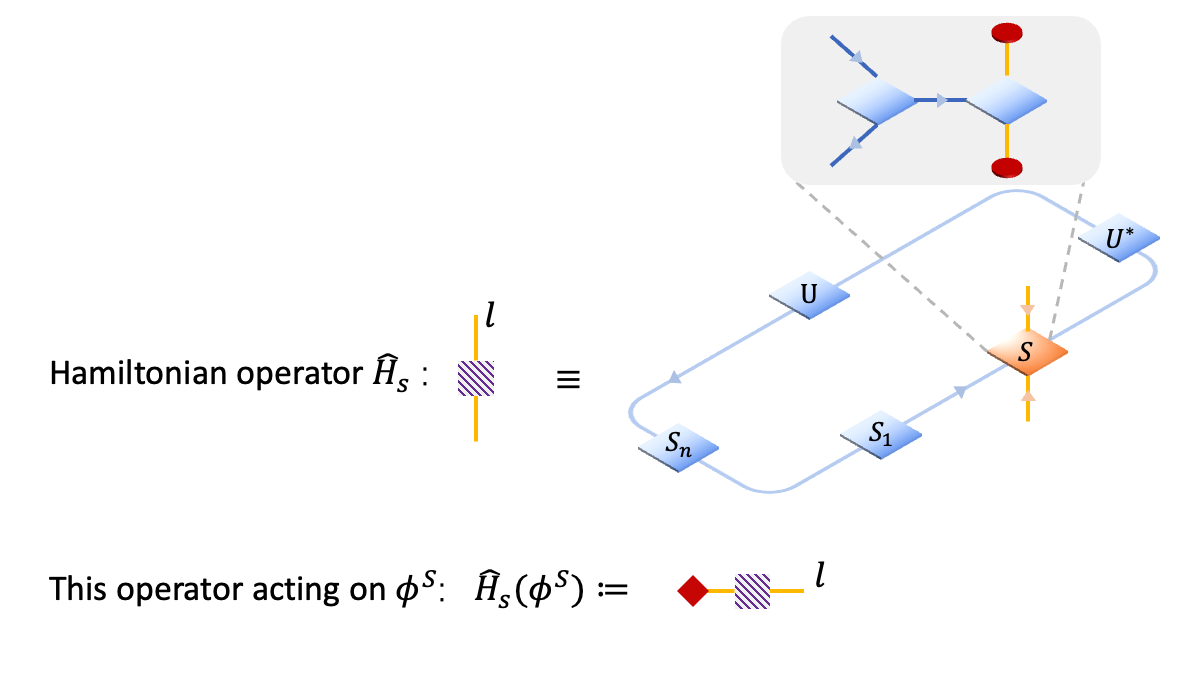}
\vspace{-4ex}
\caption{
The second step of spatial mixing: constructing the effective Hamiltonian operator by contracting the embeddings of $\phi^s$ neighbor nodes with their matrix product state kernels. This process is inspired by the sweeping procedure of DMRG, and the resulting operator acts on $\phi^s$ for updating the representation. We note that the matrix-kernel of each node itself follows a matrix-product structure. For a comparison with a similar figure, see Figure \ref{fig:DMRG} in the appendix.
}
\label{fig:spatial_mixing2}
\vspace{1ex}
\end{figure}

\subsection{An Equivariant Modification of Tensor Networks}
The application of the spatial aggregation module to geometric graphs introduces two key challenges that need to be addressed: 1. Ensuring that the spatial aggregation remains invariant under the permutation of nodes; 2. Assuring that the spatial aggregation remains equivariant when the input embeddings are subjected to rotation.

Recalling the standard pipeline for mapping raw data to a tensor network (\cite{stoudenmire2016supervised, glasser2018supervised}) (see also Table \ref{table:1}), the first \textbf{feature mapping} step involves lifting the scalar-valued raw data, typically of low dimension, into a high-dimensional complex-valued vector space. Formally, if we denote the scalar raw data as $x$, the mapped data $\varphi_x$ can be represented as:
\begin{equation} \label{eq: feature mapping}
\varphi_x := \varphi^1(x) \ve_1(x) + \cdots + \varphi^n(x) \ve_n(x).
\end{equation}
In this equation, $\ve_i$ represents a set of basis vectors, and $\varphi_x^i \in \mathbb{C}$ signifies the coefficients determined by these basis vectors and the raw data $x$. This process is akin to projecting a quantum state into a basis of a Hilbert space. When no SE(3) symmetry is involved, the basis vectors $\ve_i$ can be set as $\ve_i = (0, \ldots, 1, \ldots, 0)$, allowing the coefficients $\varphi_x^i$ to be directly fed into the tensor network. We note that a good choice of the basis and its corresponding feature mapping may be crucial for the final performance; see \citep{stoudenmire2016supervised,glasser2018supervised} for examples of feature mapping.

However, when dealing with SE(3) transformations, it is imperative to ensure that the coefficients $\varphi_x^i$ remain invariant. To achieve this, the basis vectors $\ve_i(x)$ must exhibit invariance under SE(3) transformations. This geometric invariance is characterized as follows:
\begin{equation}
x \rightarrow gx \Rightarrow \ve_i(x) \rightarrow g \ve_i(gx).
\end{equation}
For each basis vector, this transformation must hold for every $g \in SE(3)$. Notably, such invariant frames exist for each node, and an example is provided in Appendix \ref{ap: artitectures}. In the context of a geometric graph, we initially obtain three coefficients $\varphi_x^i$ for each node $x$. These three invariant coefficients can subsequently be embedded into a high-dimensional representation as usual vector embeddings.

In conclusion, we achieve $SE(3)$ equivariance by expressing the equivariant 3D coordinates by equivariant bases, which is also called \textbf{Scalarization} in \citep{du2023new}. The formulas of constructing edge- and node-wise equivariant frames out of a geometric graph are provided in (), see also (13) and (14) of \citep{du2023new}. 
\paragraph{Remark on the output} In the context of geometric graphs, the desired output may encompass invariant scalars or equivariant vector fields. Given that the input embedding of our spatial mixing module is inherently invariant, and the matrix product state is designed to transform invariant quantities into invariant quantities, the output naturally inherits this invariance property. However, if an equivariant output is desired, it can be achieved by pairing the original output with equivariant frames once again. For detailed information on this process, please refer to the vectorization block in \citep{du2022se} (formula (6)).

\paragraph{Parameterization of the spatial aggregation kernel}
Following the invariant feature mapping step \ref{eq: feature mapping}, we obtain a feature embedding $\phi(x) =\{\phi^i(x)\}_{i=1}^d$ for each node $x$. Then, we further introduce its complex conjugate $\bar{\phi}(x) =\{\bar{\phi}^i(x)\}_{i=1}^d$ as the input of the spatial aggregation kernel. Now we consider the aggregation from the node $i$'s neighbors indexed by $j$ to $i$ itself. Fixing an edge $e_{ij}$, the edge-wise spatial aggregation kernel is parameterized as a matrix product operator with one virtual leg and four physical legs: 
\begin{equation} \label{eq: kernel}
K_{abmn}(e_{ij}) = \sum_{\sigma} G_{ab}^{\sigma} \cdot A^{\sigma}_{mn}(e_{ij}), \end{equation}
where $\sigma$ denotes the virtual dimension within each aggregation kernel, and is usually taken to be small for a better efficient-expressiveness trade-off. 
Note that index $(a,b)$ will be contracted with the embedding $(\phi(x), \bar{\phi}(x))$, and the $A(e_{ij})$ is paramterized by a standard Hypernet \citep{ha2017hypernetworks} with respect to the edge feature $e_{ij}$. In this paper, we will apply a specific frame transition feature (Appendix \ref{ap: artitectures}) as the default edge feature. Then, the ``renormalization'' direction is along the index $(m,n)$: 
\begin{equation}  \label{eq: R_mn}
R_{m_0n_{N-1}}(i) = \sum_{n_0,\dots,m_{N-1}} [\phi^a(0) K_{abm_0n_0}(e_{i0}) \bar{\phi}^b(0)]\cdots[\phi^a(j) K_{abm_jn_j}(e_{ij}) \bar{\phi}^b(j)]\cdots ,\end{equation}
where we denote the virtual dimension of $m_i,n_i$ by $\chi$.
From the formula, the two-rank tensor $R_{m_0n_{N-1}}(i)$ aggregates the information from the neighbor of node $i$. Finally, the effective Hamiltonian operator $\hat{H}_i$ with respect to $i$ is formulated as:
\begin{equation} \label{block:sptial}
\hat{H}_{ab}(i) = \sum_{m,n} R_{mn} S_{ab}^{nm},  
\end{equation}
where $S$ is a node kernel that may depend on node $i$ and the layer index $t$. Now, we obtain the MPS-based aggregation formula that replaces Equation \ref{eq: mpnn operator}:
\begin{equation} \label{eq: mps mpnn}
x_i^t = \hat{H}_i (x_i^{t-1}) = \sum_b \hat{H}_{ab}(i) (\phi^b(x_i^{t-1})).  
\end{equation}
In conclusion,  besides $\{A(e_{ij}\}$ which is parameterized by a neural network, $\{R_{mn}\}$, $\{G_{ab}^{\sigma}\}$ are also tensor parameters. 

\paragraph{Permutation symmetry}  
In the context of a general graph $(V,E)$ with $|v|=N$ nodes, there is no inherent natural ordering between nodes, that is the mapping $f$ from the input to the output satisfies:
$$f(\dots, x_i, \dots, x_j, \dots) = f (\dots, x_j, \dots, x_i, \dots),\ \ \text{for}\ \ \forall 1 < i,j <N.$$
This permutation symmetry complicates the parameterization of the matrix product state via induction from 1 to N for the spatial mixing module. However, geometric graphs offer an advantage: they possess an intrinsic order for a given node's neighbors based on Euclidean distances (which also fits the locality anstz in quantum many body systems). Specifically, for nodes $x$ and $y$ within the neighborhood of node $z$ ($x, y \in N(z)$), we can establish an order such that $x$ precedes $y$ if $d(x, z) < d(y, z)$.
In other words, we assign a sequential order on neighborhoods according to a distance function, as illustrated in Figure \ref{fig:spatial_mixing1}.

Nevertheless, a challenge arises when multiple points lie within a ball of the same radius around node $z$. In such cases, where $d(x, z) = d(y, z)$, the principle of permutation symmetry requires that the contraction of $x$ and $y$ with the matrix product kernel remains commutable:
\[
\varphi_x^m \varphi_y^{m'} K^{mn}_{ab}(e_{zx}) K^{m'n'}_{bc}(e_{zy}) \bar{\varphi}_x^n \bar{\varphi}_y^{n'} = \varphi_x^m \varphi_y^{m'} K^{m'n'}_{bc}(e_{zy}) K^{mn}_{ab}(e_{zx}) \bar{\varphi}_x^n \bar{\varphi}_y^{n'}.
\]
A sufficient condition for this commutativity constraint to hold is that the matrices $M^{\sigma}_{ab}(x)$ and $M^{\sigma'}_{bc}(y)$ inside Equation~\ref{eq: kernel} are diagonalizable under the \textbf{same} unitary transformation for all values of $\sigma$:
\begin{equation} \label{eq: diagonize}
G^{\sigma}_{ab}(x) = U \text{diag}(x) U^* \quad \text{and} \quad G^{\sigma'}_{bc}(y) = U \text{diag}(y) U^*.
\end{equation} 
Remarkably, this condition is both sufficient and necessary when $K^{mn}_{ab}(x)$ and $K^{m'n'}_{bc}(y)$ are parameterized by normal matrices. Furthermore, the commutativity property implies permutation symmetry, as the permutation group can be generated by transpositions. Combining Equations~\ref{eq: mps mpnn} and \ref{eq: diagonize}, we provide a visual illustration of our equivariant MPS-based message passing in Figure \ref{fig:spatial_mixing2}.

\begin{figure}[tb!]
\centering
\includegraphics[width=0.8\textwidth]{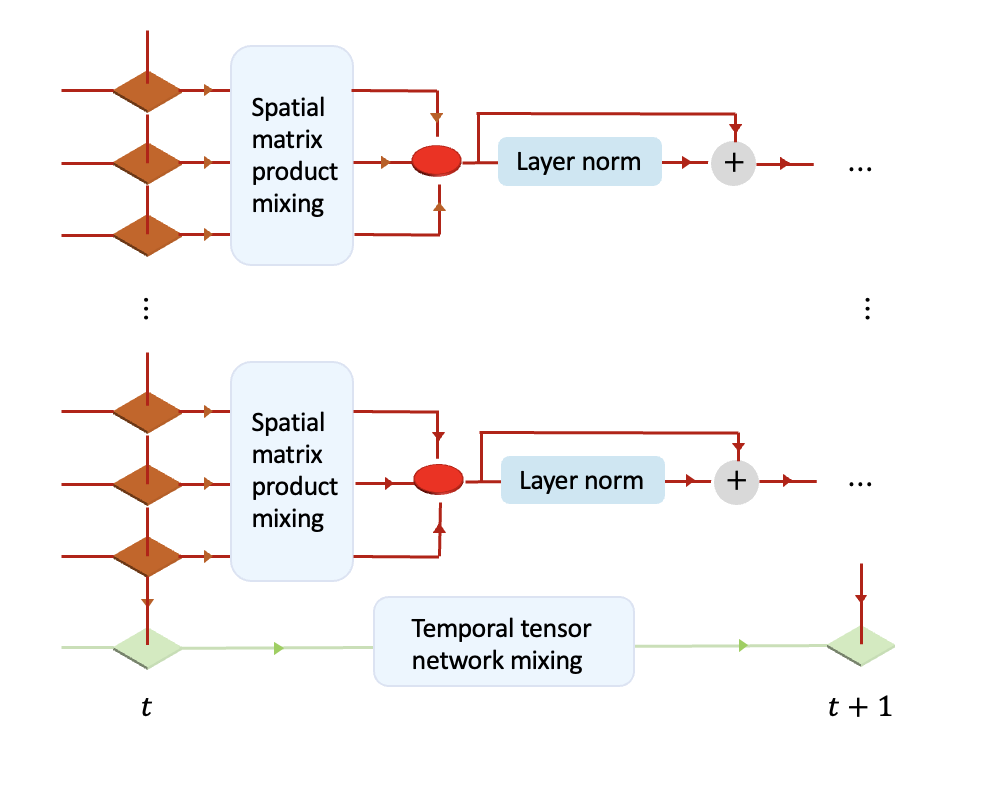}
\vspace{-4ex}
\caption{
Stacking each spatial and temporal layer will form our quantum inspired deep neural network architecture. We also add the standard residual and layer-normalization blocks as a node update between each layer \citep{vaswani2023attention}.
}
\label{fig:artitecture}
\vspace{-1ex}
\end{figure}
\subsection{Deep Architectures}
In the previous section, we introduced a novel spatial aggregation method \Model{}  based on Matrix Product States (MPS). This method serves as an extension of the traditional Message Passing Neural Network (MPNN), as described in Equation \ref{eq: mpnn operator}, from one layer ($t-1$) to the next layer ($t$). By stacking multiple layers of matrix product spatial and temporal mixing modules, we can construct a deep Matrix Product Neural Network for geometric graphs. We provide a general visualization of our deep architecture in Figure \ref{fig:artitecture}.  Notably, the node update layer and the final output layer remain to be determined. As demonstrated in the previous section, the output of our matrix product layers exhibits $SE(3)$ invariance. On the other hand, the original node update layer and the final output layer are ready to be modified for incorporating invariant and equivariant quantities such as vectors and higher order tensors. This flexibility opens the door to integrating our temporal and spatial modules into other equivariant neural networks.

\paragraph{Scalarization}
It is important to note that the only requirement for incorporating our method \Model{} is that the input of both the spatial and temporal mixing blocks should be $SE(3)$ invariant. This condition is automatically satisfied for invariant message-passing graph neural networks, such as \citep{schutt2018schnet}, \citep{unke2021spookynet}. However, for architectures that involve equivariant information, as seen in \citep{jing2020learning}, \citep{satorras2021n}, and \citep{du2023new}, we employ the scalarization method presented in \citep{du2022se} to transform the equivariant components of the representation.

For instance, the tensor product of spherical harmonics can be scalarized using techniques like CG-decomposition, while the tensor product of vectors can be scalarized by projecting them onto the node-equivariant frames (see Appendix \ref{ap: artitectures} for details). As we have demonstrated in the previous section, this scalarization procedure can be seen as analogous to expressing a quantum state in a physical basis, followed by feeding the coefficients of the expansion into a tensor network.

To obtain an equivariant output from our matrix product block, we implement the reverse process, known as tensorization, as presented in Equation 6 of \citep{du2022se}. This allows us to recover equivariant information from the transformed representations, ensuring compatibility with architectures that involve both invariant and equivariant features.
For example, let $\mathcal{F}$ represent an equivariant operator, and $\mathbf{O}$ denote the output of our combined temporal and spatial modules. The integration into an equivariant neural network can be mathematically expressed as:
\begin{equation} \label{method:  merge}
\dots \xrightarrow{\text{Spatial MPS}}\mathbf{O} \xrightarrow{\text{Tensorization}}  \mathcal{T}(\mathbf{O}) \xrightarrow{\text{Equivariant Layer}} \mathcal{F}(\mathcal{T}(\mathbf{O})) \xrightarrow{\text{Scalarization}} 
\mathcal{S}\mathcal{F}(\mathcal{T}(\mathbf{O}))\ldots, \end{equation}
where $\mathcal{S}$ and $\mathcal{T}$ represent the scalarization and tensorization procedure, respectively. The formal formulas of $\mathcal{S}$ and $\mathcal{T}$ are provided in Appendix \ref{ap: artitectures}.
This approach allows us to preserve the equivariant nature of the network while benefiting from the spatial aggregation and temporal mixing capabilities offered by our quantum-inspired matrix product layers. The flexibility of our approach and its compatibility with existing equivariant neural network structures make it a promising candidate for a wide range of applications.

\subsection{Architecture Merging}
\label{merging}
Our tensor network based spatial and temporal aggregation method \Model{} is flexible for merging into other deep architectures that can model euqivariant quantities like vector fields and Hamiltonian. 

\paragraph{SpaTea\_Clofnet.}
In this study, we explore the integration of our spatial and temporal mixing layers in Equation~\ref{eq: temporal mps} with an equivariant final output layer, using the classical Clofnet~\citep{du2022se} as our example. We replace the attention mechanism with edge features used in Clofnet's MPNN phase with our matrix-product spatial mixing module in Equation~\ref{block:sptial}. Given that the node representation in Clofnet is purely invariant, the temporal mixing component can be applied without any modifications. Next, we inherit the equivariant output layer from Clofnet by pairing the invariant output with edge frames, ensuring that equivariant features are appropriately maintained in the final layers of the network. 

\paragraph{SpaTea\_DeepH-E3.}
DeepH-E3 \citep{Gong_2023} is invented as an advanced deep neural artiteture for predicting the DFT-Hamiltonian. Different from equivariant vector fields, DFT-Hamiltonian tansforms according to higher dimensional group representations of $SO(3)/O(3)$. 
DeepH-E3 incorporates both invariant and equivariant components in its message-passing phase and edge update layers. We replace the invariant message-passing of DeepH-E3 by Equation~\ref{block:sptial} directly. The edge-update layers with the Clebsch–Gordan coefficients are left unchanged to make sure that our method's output respects Hamiltonian's symmetry. 

We note that detailed formal formulas for both SpaTea\_Clofnet and SpaTea\_DeepH-E3 are provided in Appendix \ref{ap: artitectures}.

\subsection{Discussion}
From our detailed construction of spatial and temporal aggregation method, the matrix product consists of polynomials for each component of the node-wise vector embedding $\phi^a(j)$, where index $j$ is ranged from the neighbors of a given node.
Suppose the initial feature of node $i$ is a scalar $x_i$, and $\phi^a(j)$ consists of monomials: $\phi^a(j): = \{1, x_i,\dots, x_i^n\}$, then we can easily conclude that for an infinite virtual dimension $\chi$ of formula \ref{eq: R_mn}, the spatial temporal aggregation is able to express many-body polynomials up to an arbitrary order. The similar conclusion holds for the temporal aggregation, thus lead to a more powerful READ-OUT function of different layers than traditional layer aggregations. On the other hand, we usually perform dimension cutoff for the virtual index $\chi$ following the classical tensor network algorithms for modeling quantum systems \citep{banuls2023tensor}, which leads to the following additional privileges. 

\paragraph{Quantum privilege}
From a quantum computing perspective, our algorithm's adaptability for implementation as a classical-quantum hybrid algorithm \citep{10059764} is significant. It opens the door to utilizing the advantages of quantum computation in combination with the computational efficiency and easy compression of our quantum-inspired approach.
Furthermore, the original invention of Matrix Product State (operator) aimed to efficiently compress many-body quantum states. Consequently, our quantum-inspired algorithm is expected to inherit two valuable traits from the quantum physics perspective: 

1. \textbf{Ease of Compression}: Like its quantum counterpart, our algorithm allows for efficient compression by cutting off the dimension of $\chi$ for each layer, reducing the computational complexity and memory requirements for distilling our MPS-based large pre-trained model.

2. \textbf{Quantum-Compatible}: The classical-quantum hybrid algorithm in a quantum computer setting can also be implemented, taking advantage of the quantum-inspired nature of our approach.

To create a parameterized version of the tensor network for efficient compression, we must parameterize our spatial and temporal matrix product states in a canonical form, specifically the central-orthogonal form. The five necessary conditions for the central-orthogonal form are elaborated in the appendix. It is essential to understand that all tensors can be represented in a central-orthogonal form through tensor-train decomposition. This means that not only our matrix-product state-based algorithm but also all tensorized neural networks can be transformed into this canonical form. However, this transformation process involves flattening $N-1$ dimensions' indices and performing the QR decomposition \citep{wendland_2017}.

It's worth mentioning that ordinary neural networks can also be compressed by tensor networks by reshaping the linear layers, see \cite{Gao2019CompressingDN, LIU2022108171}.   On the other hand, since our modules possess an explicit matrix-product structure, we can obtain their canonical form by individually conducting local Singular Value Decomposition (SVD) for each kernel $k(e_{ij})$. The detail is left in Appendix \ref{eq: canonical decompostion}.

\section{Experiment}
To underscore the effectiveness of our geometric graph tensor network, we conducted comprehensive assessments across three intricate many-body system prediction tasks, encompassing both dynamical state prediction and property prediction. Given the inherent entanglement within these systems, we assert that the utilization of tensor network-based spatial and temporal aggregations, as explicated in our methodology section, surpasses the efficacy of conventional geometric Graph Neural Networks (GNNs). This approach allows us to address two distinct types of geometric symmetries: many-body  dynamcis and quantum tensors. 
\subsection{Newton Dynamics Prediction}
In this experiment, we employ the {SpaTea\_Clofnet}, as detailed in Section~\ref{merging}, to predict future positions of synthetic many-body systems driven by Newtonian force fields. This task is inherently equivariant, as any rotations or translations applied to the initial system state yield identical transformations on the future system positions.

Following the setup proposed in \citep{fuchs2020se3transformers, du2022se}, we consider a 20-body charged system controlled by pre-designed Newtonian force fields. We present three intricate force fields following \citep{du2022se} (formulas in (44), (45), and (46)), despite their two-body nature. Our goal is to forecast the future positions of particles in these many-body systems, which exhibit complex entanglement. Consequently, we anticipate that our {SpaTea\_Clofnet} outperforms Clofnet in these scenarios.


\paragraph{Dataset}   The three systems under evaluation each consist of 20 nodes (particles) and are influenced by electrostatic force fields (denoted as ES(20)), an additional gravity field (denoted as G+ES(20)), and a Lorentz-like force field (denoted as L+ES(20)). We generated these SE(3)-equivariant datasets using the descriptions and source code from \citep{du2022se}, with each dataset comprising 7,000 total trajectories. Results are presented in terms of mean squared error (MSE) between a method’s node position predictions on the test dataset and the corresponding ground-truth node positions after 1,000 timesteps. Additionally, results for a 40-body system are provided in the appendix.

\begin{table*}[t]
  \caption{MSE for future position prediction over four datasets. }
  \label{table:nbody}
  \centering
  \begin{tabular}{l ccccc}
    \toprule
      Model  & ES(20) & G+ES(20) & L+ES(20) \\
    \midrule
    GNN  & 0.0720	& 0.0721 & 0.0908 \\
    TFN & 0.0794& 0.0845 & 0.1243 \\
    SE3 Transformer  & 0.1349 & 0.1000& 0.1438 \\
    Radial Field  & 0.0377 & 0.0399 & 0.0779 \\
    EGNN  & 0.0128 & 0.0118 & 0.0368 \\
    GCPNet & 0.0077 &0.0001 &0.0172 \\
    ClofNet  & 0.0079 & 0.009 & 0.0230\\
    \midrule
    SpaTea\_Clofnet & 0.0076 &0.0076 & 0.0189\\
    \bottomrule
  \end{tabular}
  \vskip -0.1in
\end{table*}


\paragraph{Results} Besides the standard baselines provided in \citep{du2022se}, we add a new equivariant model GCPNet following \citep{morehead2022geometry}. The results are provided in Table \ref{table:nbody}. As we can see from the table, {SpaTea\_Clofnet} achieves supreme results for all the three dynamical scenarios, indicating that our spatial and temporal aggregations based on tensor-network are effective. For example, in the cases of G+ES(20) and L+ES(20), our method enhances the performance of the state-of-the-art approach ClofNet, reducing the error from 0.009 to 0.0076 and from 0.023 to 0.0189, respectively.
These improvements correspond to  a notable reduction in errors of 15\% and 17\% for the two cases.

\subsection{Quantum Tensor Prediction}
In this section, we apply the {SpaTea\_DeepH-E3} as discussed in Section~\ref{merging}, which is modified from the DeepH-E3 model in \citep{Gong_2023} for predicting the Hamiltonian tensor of density functional theory $H_{DFT}$.  We provide the detail of how to incorporate our spatial and temporal matrix-product method into the DeepH-E3 model in \citep{Gong_2023} in Appendix \ref{ap: artitectures}. In a nutshell, $H_{DFT}$ is an equivariant physical tensor that transforms according to the representation theory of $O(3)$. More precisely, $H_{DFT} = \{H_{i p_{1}, j p_{2}}\}_{m_1m_2}^{l_1l_2}$, where $i,j$ are the body index,  p is the multiplicity index, and $l,m$ are the angular momentum quantum number and its corresponding magnetic quantum number (which is necessary for identifying a spherical harmonic function).
For the reader's convenience, we copy the transformation rule of $H_{DFT}$:
\begin{equation}
    \left(H_{i p_{1}, j p_{2}}^{\prime}\right)_{m_{1} m_{2}}^{l_{1} l_{2}}=\sum_{m_{1}^{\prime}=-l_{1}}^{l_{1}} \sum_{m_{2}^{\prime}=-l_{2}}^{l_{2}} 
    D_{m_{1} m_{1}^{\prime}}^{l_{1}}(\mathbf{R}) D_{m_{2} m_{2}^{\prime}}^{l_{2}}(\mathbf{R})^{*}\left(H_{i p_{1}, j p_{2}}\right)_{m_{1}^{\prime} m_{2}^{\prime}}^{l_{1} l_{2}} ,
    \label{eq:transform_H}
\end{equation}
where $\mathbf{R}$ is any rotation matrix and $D^l_{mm'}(\mathbf{R})$ is the corresponding Wigner D-matrix.

\paragraph{Dataset} We test our model's performance following the setting in \cite{Li_2022,Gong_2023}. The datasets are comprised of DFT supercell calculation results of MoS2, and different geometric configurations are sampled from ab initio molecular dynamics. Five hundred structures with 5 × 5 supercells are generated by ab initio molecular dynamics performed by VASP with PAW pseudopotential
and PBE functional. 
\paragraph{Results} The test results
are summarized in Table \ref{table:dft} and compared with those of the original
DeepH and DeepH-E3.  Our experiments
show that the mean absolute errors (MAEs) of Hamiltonian
matrix elements averaged over atom pairs are all within a fraction of a
meV. We rerun the DeepH-E3 experiment with the same hyper-parameters as \cite{Gong_2023} for 1000 epochs with Adam optimizer and ReduceLROnPlateau as our learning rate scheduler. As we can see from the table, 
our method consistently delivers superior results across all cases, showcasing a substantial margin of improvement.
For example,Our strategy outperforms the DeepH-E3 method by 25\%, 33\%, 32\%, and 36\% in the Mo-Mo, Mo-S, S-Mo, and S-S cases, respectively. Furthermore, our method exhibits significant improvements over DeepH, with enhancements of 63\%, 62\%, 52\%, and 58\% for the Mo-Mo, Mo-S, S-Mo, and S-S orbitals, respectively.

\begin{table*}[t]
  \caption{MAE for DFT Hamiltonian matrix elements averaged
over atom-pairs of MoS2. Unit: $mev$.}
  \label{table:dft}
  \centering
  \begin{tabular}{l ccccc}
    \toprule
      Model  & Mo-Mo & Mo-S & S-Mo & S-S \\
    \midrule
    DeepH  & 1.3 & 1.0 & 0.8 & 0.7 \\
    DeepH-E3 & 0.63 &0.57 &0.56 &0.46 \\
    \midrule
    SpaTea\_DeepH-E3  &0.47 & 0.38 & 0.38 & 0.29  \\
    \bottomrule
  \end{tabular}
  \vskip -0.1in
\end{table*}

\section{Related Work}
We here extend the discussions on  related works presented in the Introduction and Preliminaries sections. 

The intersection of tensor networks and deep learning has witnessed significant developments in recent years. For instance, modeling sequential data, such as time series and natural language, as a long-ranged tensor has demonstrated its effectiveness in uncovering intricate correlations among sequential tokens 
\citep{tjandra2016gated, goessmann2020tensor, pestun2017tensor,huang2017tensor}. Notably, some research has adopted the spatial and temporal terminologies for modeling space-time scenarios \citep{zhou2019spatial, xu2022spatial}, aligning with how traditional Convolutional Neural Networks (CNN)  \citep{o2015introduction}), and Recurrent Neural Networks (RNN), including Long Short-Term Memory (LSTM)  \citep{sherstinsky2020fundamentals}, approach these aspects. CNN treats spatial information as regular regions in a grid mesh, a contrast to our irregular graph-based setting. On the other hand, RNN interprets temporal aspects as the sequential order of data in a sequence. In fact, when stacking an infinite number of layers with residual connections, the limit of this process becomes a temporal Ordinary Differential Equation (ODE), as seen in Neural ODE \citep{chen2018neural}. On the other hand, the spatial and temporal design is intrinsically embedded in our model, regardless of the data types.

In our research, we introduce the Matrix-Product-State as an effective approach for tackling the tensor modeling required by the Geometric Universality Theorem in \citep{du2022se}, while circumventing the exponential wall issue. We provide a concrete example of how many-body entangled geometric information affects the geometric expressiveness in appendix \ref{ap: artitectures}. Besides geometric graph data, other modalities of data like natural language also reveals the entangled property, see \cite{10.1145/3308558.3313516, zhang2019generalized}.
For a theoretical understanding, the reader can consult the tensorized tangent kernel theory \citep{guo2021neural} for measuring the expressiveness of 1-dimensional tensor network in the infinite regime. 
Analogous to the impossibility of increasing the width of a neural network to infinity, the virtual index of our parameterized Matrix-Product-State also remains finite. Consequently, understanding the expressiveness of this finite setting is a challenging problem. It is worth noting that exploring entanglement entropy, which can be easily calculated if we parameterize the Matrix-Product-State in the canonical form, and its relationship with the area law \citep{decker2022many}, can provide insights into the expressiveness of various tensor networks and autodecoders based on tensor networks \citep{eisert2010colloquium}. 

\paragraph{Complex-Valued Neural Network}
After feature mapping, all tensors are transformed into complex-valued tensors, and the transformation map between tensors is also complex-valued. However, tensor networks and learning algorithms based on quantum computing are not the sole method for constructing complex-valued transformations. Architectures like Fourier Neural Operator (FNO) also employ complex-valued Fourier transforms \citep{li2020fourier}, and \citet{rajesh2021quantum} introduced the imaginary part into the Vision Transformer to convey the quantum ``phase" information between pixels.

\paragraph{Equivariant Graph Neural Network}
Recall that, our spatial and temporal modules based on Matrix-Product-State can seamlessly integrate into prevalent equivariant neural networks. At a high level, our modules can replace any equivariant aggregation, provided the input remains invariant. This adaptation extends to models like Schnet \citep{schutt2021equivariant}, Spookey-net \citep{unke2021spookynet}, and others. For a comprehensive categorization, the readers can consult \citep{
liu2023symmetry}. Beyond invariant aggregation, an alternative research avenue leverages spherical harmonics as equivariant embeddings and transformations. In this paper, the tensor network's input comprises the invariant coefficients of an equivariant quantum state, expanded through an equivariant basis. Besides our invariant feature mapping approach, another common approach for implementing non-parameterized tensor networks in physics systems with gauge symmetry is to directly input the equivariant quantum state. However, this approach constrains the tensor network's design, as it must be SE(3) equivariant, exemplified by \citep{weichselbaum2012non, slagle2023quantum}.

\section{Conclusion and Outlook}
We introduced \Model{}, a novel equivariant Matrix Product State (MPS)-based message-passing strategy. \Model{} utilizes an efficient implementation of tensor contraction to effectively model intricate many-body relationships, avoiding mean-field approximations and capturing symmetries within geometric graphs. 
 \Model{}  also enables a straightforward substitution of the standard message-passing and layer-aggregation modules inherent in geometric GNNs, requiring minimal effort. Through empirical verification, we demonstrated the superior performance of \Model{} in predicting classical Newton systems and quantum tensor Hamiltonian matrices. 

We posit that \Model{}'s adaptability to existing GNNs and  modeling capabilities beyond mean-filed approxmiation make it a versatile tool applicable across diverse fields, including materials science, chemistry, physics, drug discovery,  quantum computing, and beyond. 
For example, 
the definition of the Matrix-Product-State underscores that matrix product operations establish entanglement between individual tensor states. Notably, the entanglement procedure is vital in quantum computing and is realized through quantum circuits. Consequently, effectively parameterized tensor network architectures serve as a guiding principle for shaping parameterized quantum circuits, as detailed in \citep{huggins2019towards}. Reciprocally, every quantum circuit can be represented as a tensor network, with the bond dimension contingent on the circuit's width and connectivity. The quantum tangent kernel method \citep{shirai2021quantum} also leverages these concepts. 
We detail the possible extensions of our method along this direction in Appendix \ref{ap: extentions}.

\bibliography{iclr2024_conference}
\bibliographystyle{iclr2024_conference}

\appendix
\section{Appendix}

\subsection{DMRG for tensor network contraction} \label{AP: DMRG}
\begin{figure}[tb!]
\centering
\includegraphics[width=0.8\textwidth]{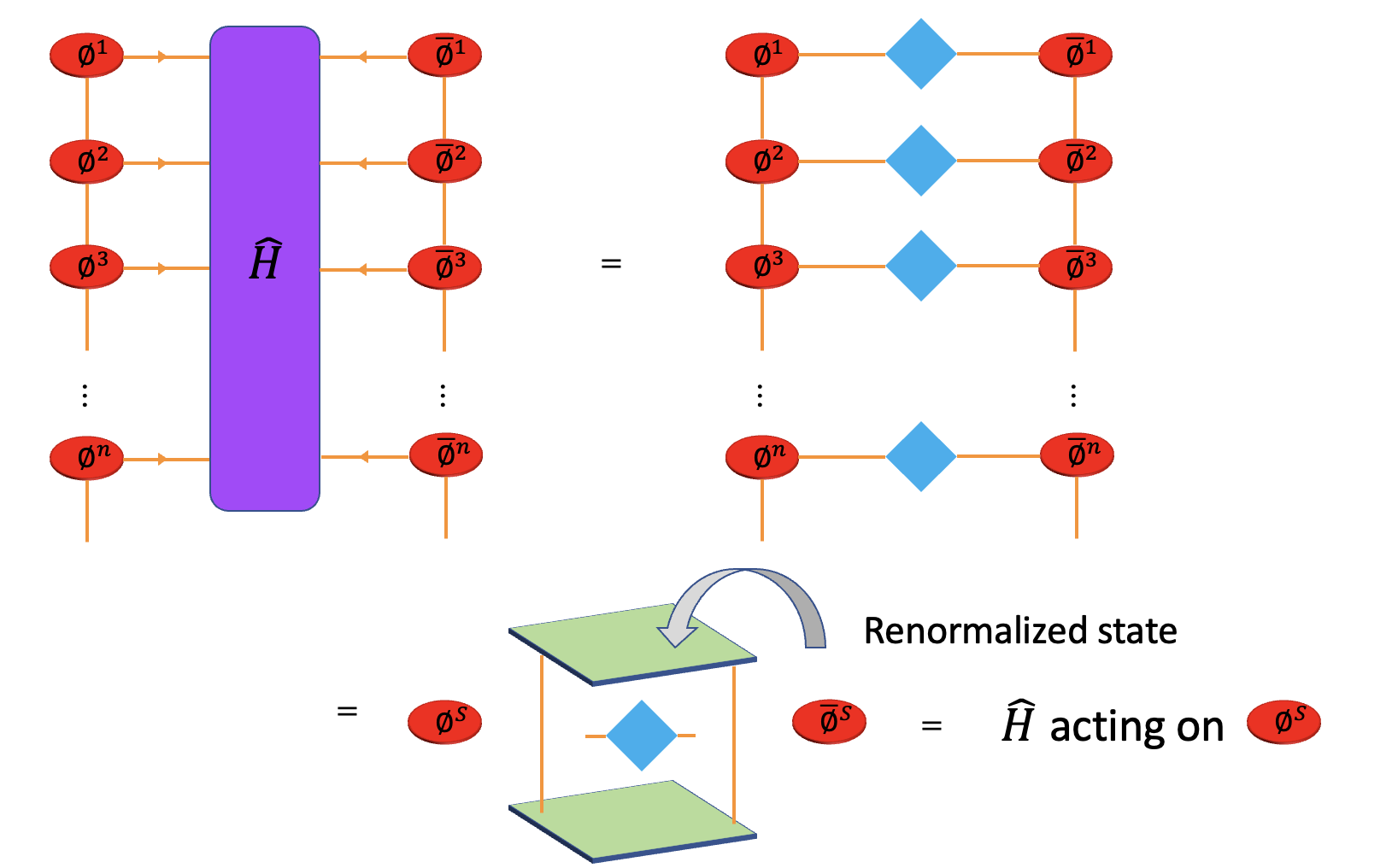}
\vspace{1ex}
\caption{
The renormalization in DMRG refers to the fact that the effective Hamiltonian for each node renormalizes (by contraction) the information from other nodes, as shown in this figure. Here, the effective Hamiltonian $\hat{H}_{\text{effective}}$ acting on the s-th state consists of the green renormalized states and the blue operator acting solely on state $s$. 
}
\label{fig:DMRG}
\vspace{-1ex}
\end{figure}

Our spatial tensor-network module is inspired by DMRG \citep{white1993density,hyatt2019dmrg}, which is invented for calculating the ground state energy of a Hamiltonian system $\hat{H}$:
$$E : = \min_{|s\rangle}  \langle s | \hat{H} | s \rangle\ \ \text{s.t.}\  \langle s | s \rangle = 1.$$
We note that this is essentially a tensor contraction problem and the visualization is given by the left hand side of \ref{fig:spatial_mixing2}. However, for a N-qubit system, $| s \rangle$ lives in a $2^N$ dimensional vector space, which makes this global combinatorial optimization problem impossible. DMRG solves this problem by transferring the global Hamiltonian $\hat{H}$ to node-wise effective Hamiltonian. Then, at each step of DMRG, the optimization is only down for a single state $|s_i\rangle$, which is doable. Then, we sweep along each node back and forth until the final procedure converge. We provide a simplified example in Figure \ref{fig:DMRG}, where the many-body Hamiltonian $\hat{H}$ is already in the matrix-product form. Then by contracting all other states excepts $\phi^s$, the energy $E$ is given by
$$E = \hat{H}_{\text{effective}}(\phi^s, \Bar{\phi}^s),$$
as shown in the last line of Figure \ref{fig:DMRG}.

\subsection{Possible Extentions} \label{ap: extentions}
Our spatial and temporal aggregation kernels utilized 1-dimensional tensor network (matrix product) as the backbone parameterization method. However, besides 1-dimensional tensor network. there exists higher dimensional tensor networks with more complex topology \citep{verstraete2004renormalization, Pan_2020, xu2023graph}. It is worth noting, nevertheless, that such tensor networks are usually designed for specific physics problems, and such ansatz \citep{Felser_2021} may not exist for general geometric graphs. An alternative approach is to utilize automatic architecture search methods for tensor networks, see \cite{Li2020EvolutionaryTS, 10.5555/3618408.3619249}.
\paragraph{Hybrid classical-quantum algorithm}
Our spatial and temporal mixing blocks can be modified to adapt the quantum algorithm scheme. Roughly speaking, parameterized quantum algorithm is composed of two components: 
\begin{enumerate}
    \item \textbf{Quantum Gates}: quantum gates transform a quantum state to another quantum state by unitary transformations. Therefore, one-quobit quantum gates (e.g., Pauli gates \citep{nielsen00}) are similar with the linear fully-connected layers of classical neural networks. On the other hand, there are also many-quobit quantum gates (e.g., CNOT gates \citep{nielsen00}), which play a similar role as convolution layers of classical neural networks. In fact, any operation possible on a quantum computer can be reduced to a set of basic gates. For example, the rotation operators, the phase shift gate, and the CNOT together form a set of universal quantum gates.
    \item \textbf{Quantum Circuits}: quantum circuits are the organized collection of quantum gates. As the universal quantum gates are finite, the design space of quantum algorithms is exactly the space of all possible combinations of these universal quantum gates.
\end{enumerate}
Therefore, to make our algorithm quantum, we can simply replace the multi-layer fully connected forward layers by rotation operators acting on each node. On the other hand, our spatial mixing and temporal mixing modules are composed of matrix-product states, which can be realized by quantum circuits following the standard pipeline in \cite{Huggins_2019, pan2023efficient}. 

\paragraph{Compression}

Consider the $K_{abmn}(e_{ij})$ defined by Equation~\ref{eq: kernel}, and we merge indices $a$ and $b$ contracting with the state embeddings into a single index $\sigma := (a,b)$. Then, the renormlized information has the following matrix-product form:
\begin{equation} \label{eq: renormal of mps}
R^{\sigma}(i) = \sum_{\sigma_1, \dots , \sigma_N} K^{\sigma_1} \cdots K^{\sigma_j} \cdots K^{\sigma_N}.    
\end{equation}
Explicitly, we group the two indexes of $K^{\sigma_j}_{mn}$ together to obtain an intermediate matrix $K_{(\sigma_j, m),n}$. Then, An SVD of $K$ yields $K(j) = U_j S_j V_j$ for each neighbor $j$. Substitution of the decomposed expression into \ref{eq: renormal of mps}, we derive the canonical form of $R^{\sigma}(i)$ step by step:
\begin{equation} \label{eq: canonical decompostion}
R^{\sigma}(i) = \sum_{\sigma_1, \dots , \sigma_N} \sum_{m_1,\dots,m_N} \sum_{s_1} U_{(\sigma_0,m_0),s_1} S_{s_1,s_1}V_{s_1,m_1}K^{\sigma_1}_{m_1,m_2}\cdots.    
\end{equation}
Reshape $U_{(\sigma_0,m_0),s_1}$ to $U^{\sigma_0}$, we have
$$\sum_{\sigma_0} U^{\sigma_0} = I,$$
where $I$ denotes the identity matrix. Note that although every high order tensor admits a canonical form, our matrix-product parameterization reduces the computational complexity from performing SVD on order $O(d^N)$ matrices to order $O(N \cdot \chi \cdot d)$ matrices. For example, the formal way of transforming a r-order tensor $T_{1,\dots,r}$ is to flatten the indexes of $(1,\dots, r)$ to two parts $(u,v)$, and do SVD decomposition on $T_{uv}$. Comparing with our local matrix, the range of $(u,v)$ is of exponential order $d^{N-1}$ with respect to the body numbers.

The central-orthogonal form offers the advantage of direct compression by truncating each virtual index according to the singular values obtained from each SVD decomposition \citep{doi:10.1137/07070111X}. It has been established in \cite{oseledets2011tensor} that this truncation procedure represents the optimal low-rank reduction, ensuring efficient compression. relevant references on distillation large models by tensor network method can also be found in \cite{neill2020overview}.
Similar low-rank methods have been successfully applied in conventional neural networks to achieve cost-effective fine-tuning, as evidenced in \cite{hu2021lora}.

\subsection{Two Architectures} \label{ap: artitectures}

In the main context, we present a general procedure for stacking multiple spatial and temporal aggregations based on tensor networks (see Figure \ref{fig:artitecture}). With all geometric features scalarized, the entire architecture becomes invariant. To introduce equivariant quantities into our method, we employ both scalarization and tensorization, as illustrated in Section \ref{method:  merge}.

\subsubsection{Many-Body Entanglement through Relative Orientation}
We assume that the 3D graph's mass has been translated to zero to ensure the translation invariance of the system by placing the center of mass at the origin.

Following \cite{du2023new}, consider body $i$ and one of its neighbors $j$ with positions $\textbf{x}_i$ and $\textbf{x}_j$, respectively. The orthonormal \textbf{Edge-wise frame} $\mathcal{F}_{ij} := (\textbf{e}^{ij}_1, \textbf{e}^{ij}_2, \textbf{e}^{ij}_3)$ is defined with respect to $\textbf{x}_i$ and $\textbf{x}_j$ as follows:
\begin{equation} 
\label{eq:edge frame}
\left(\frac{\textbf{x}_i - \textbf{x}_j}{\norm{\textbf{x}_i - \textbf{x}_j}}, \frac{\textbf{x}_i \times \textbf{x}_j}{\norm{\textbf{x}_i \times \textbf{x}_j}}, \frac{\textbf{x}_i - \textbf{x}_j}{\norm{\textbf{x}_i - \textbf{x}_j}} \times \frac{\textbf{x}_i \times \textbf{x}_j}{\norm{\textbf{x}_i \times \textbf{x}_j}}\right).
\end{equation}
We define the local geometry by the scalarized cluster around each edge using equivariant edge-frames. The general definitions of scalarization and tensorization are also provided in \cite{du2022se}. By listing all the neighbors of $\textbf{x}_i$, we define the relative orientation of $\textbf{x}_j$ and $\textbf{x}_k$ by the orthogonal matrix $\mathcal{O}_{jk} := \textbf{e}_{ij} \cdot \textbf{e}_{ik}^T$. For insights into how this quantity affects the expressiveness power of graph neural networks, refer to the frame transition section of \cite{du2023new}.

However, since relative orientations encode relations between edges, we need to efficiently encode these features through a node-wise message-passing scheme. For that purpose, we introduce the node-wise equivariant frames:

\textbf{Node-wise Frame.} Let $\Bar{\textbf{x}}_i := \frac{1}{N}\sum_{\textbf{x}_j \in \mathcal{N}(\textbf{x}_i)} \textbf{x}_j$ be the center of mass around the 1-hop neighborhood of $\textbf{x}_i$. The orthonormal equivariant frame $\mathcal{F}_i := (\textbf{e}^i_1, \textbf{e}^i_2, \textbf{e}^i_3)$ is defined with respect to $\textbf{x}_i$ as follows:
\begin{equation} 
\label{eq:node frame}
\left(\frac{\textbf{x}_i - \Bar{\textbf{x}}_i}{\norm{\textbf{x}_i - \Bar{\textbf{x}}_i}}, \frac{\Bar{\textbf{x}}_i \times \textbf{x}_i}{\norm{\Bar{\textbf{x}}_i \times \textbf{x}_i}}, \frac{\textbf{x}_i - \Bar{\textbf{x}}_i}{\norm{\textbf{x}_i - \Bar{\textbf{x}}_i}} \times \frac{\Bar{\textbf{x}}_i \times \textbf{x}_i}{\norm{\Bar{\textbf{x}}_i \times \textbf{x}_i}}\right).
\end{equation}

Then, we define the orientation between an edge and a node by $$\Tilde{\mathcal{O}}(ij) := \textbf{e}_i \cdot \textbf{e}_{ij}^T. $$ Since orthogonal matrices form a group, we have $\mathcal{O}_{jk} = \Tilde{\mathcal{O}}(ij)\cdot \Tilde{\mathcal{O}}(ik)^T$. In other words, if the spatial aggregation is able to express second-order polynomials (tensors), then the graph neural network can capture the relative orientations. Similarly, the triangular orientations between three nodes require third-order polynomials.

Now, we introduce two variants of our method based on two classical equivariant graph models to demonstrate the flexibility of our approach in seamlessly integrating geometric matrix-product modules into equivariant models. From a high-level perspective, a unified method for transforming equivariant quantities into invariant scalars, essential for representing higher-order polynomials, is through \textbf{scalarization}. Given an equivariant frame $(e_1, e_2, e_3)$, an equivariant vector $\mathbf{x}$ is transformed into scalars $\mathbf{x} \rightarrow \Tilde{\mathbf{x}}: = (\mathbf{x} \cdot e_1, \mathbf{x} \cdot e_2, \mathbf{x} \cdot e_3)$. For spherical harmonics, a corresponding procedure is the CG-decomposition \citep{thomas2018tensor}.

Conversely, tensorization pairs scalars with equivariant bases, including vector frames and spherical harmonic bases. For instance, vectorization is a special case of tensorization that maps scalars $\{x^a, x^b, x^c\}$ to an equivariant vector: $\mathbf{x} = x^a e_1 + x^b e_2 + x^c e_3$.

From this viewpoint, our invariant deep architecture in Figure \ref{fig:artitecture} solely employs scalarization with invariant spatial and temporal matrix-product aggregation. In the following, we set our default invariant edge feature to be the combination of relative distances and orientations between nodes:
$$e_{ij} = (d(\mathbf{x}_i, \mathbf{x}_j) || \Bar{\mathcal{O}}_{ij}).$$
We now provide two approaches to incorporate various tensorizations into our neural network for different experimental scenarios.

\paragraph{\Model{}\_Clofnet}
Our SpaTea\_Clofnet is an adaptation of the naive message-passing block from \cite{du2022se}. In this variant, each message-passing block takes node positions $\mathbf{x}^l$, node embeddings $h^l$, SO(3)-invariant scalars $s_{ij}$ (e.g., Euler angle representation of $\Tilde{\mathcal{O}}(ij)$), and edge information $e_{ij}$ (e.g., one-hot representation of edge type) as inputs, producing transformations on $h^{l+1}$ and $x^{l+1}$. Let $\mathbf{x}_i^{0}$ represent the equivariant positions of each node. The equations for each message-passing layer are defined as follows:
\begin{gather}
    m_{ij} = \phi_m (s_{ij}, h_i^l, h_j^l, e_{ij}), \\
    h_i^{l+1} = \hat{H}_i (h_i^l), \\
    \mathcal{F}_{ij}^{l+1}=\mathbf{EquiFrame}(\mathbf{x}_i^l, \mathbf{x}_j^l), \\
    \mathbf{x}_i^{l+1} = \mathbf{x}_i^{l} + \frac{1}{N}\sum_{j \in \mathcal{N}(\mathbf{x}_i)}{\mathbf{Vectorize}(m_{ij}^l, \mathcal{F}_{ij}^l)},
\end{gather}
where $\phi_m$, $\phi_k$, and $\phi_h$ represent distinct Multi-Layer Perceptrons (MLPs) with varying parameters. It is important to note that the spatial matrix-product operator $\hat{H}_i$ for each node, as defined in Equation~\ref{block:sptial}, depends on the edge message $m_{ij}$ following Equation~\ref{eq: kernel}. In the final layer, the representation $h_i^L$ is obtained by temporally aggregating previous $h_i^l$ using the formula \ref{eq: temporal mps}:
$$h_i^L = \Psi (h_i^{0},\dots, h_i^{L-1}).$$
Through the vectorization step, the output undergoes a transformation from invariant scalars ($m_{ij}^L$) to an equivariant vector field.
 
\paragraph{\Model{}\_DeepH-E3}
Based on the graph-based deep architecture proposed in \cite{Gong_2023}, SpaTea\_DeepH-E3 outputs the equivariant Hamiltonian of a material, respecting higher-dimensional symmetries of $O(3)$ (see Equation (1) in \cite{Gong_2023}). Comprising multiple layers of node (vertex) updates and edge updates, the final edge output consists of spherical harmonics of different orders. Through the Wigner-Eckart layer, the edge feature undergoes transformation in precisely the same way as the Hamiltonian (refer to the Equivariance of the spin–orbital Hamiltonian section in \cite{Gong_2023}).

To incorporate our spatial Matrix-product aggregation method, we modify the Equiconv part of DeepH-E3. Using the notations from \cite{Gong_2023}, let $\{e_B (|\mathbf{r}_{ij}|)\}_n = \exp (-\frac{(|\mathbf{r}_{ij}|-r_n)^2}{2 \Delta^2})$ denote the Gaussian embedding of relative distances. The equivariant edge message $m_{ij}$ is aggregated as follows:
\begin{gather}
    h_i^l = h_{i0}^l \oplus h_{i1}^l \dots, \\
    s_i^{l+1} = \hat{H}_i (h_{i0}^l),\\
    v_{ij}^{l+1} = [U(s_i^l) || h_i^l )] \otimes [V Y(\mathbf{r}_{ij})],\\
    m_{ij}^{l+1} = \phi_m (e_B (|\mathbf{r}_{ij}|)) \odot \textbf{Gate}(v_{ij}^{l+1}).
\end{gather}

Here, $Y(\mathbf{r}_{ij}): = \{Y_{lm}(\mathbf{r}_{ij})\}$ represents a set of spherical harmonic bases, where $l$ denotes the angular momentum quantum number, and $m$ denotes the magnetic quantum number. For each $l$, $m$ ranges from $-l$ to $l$ ($2l+1$ dimensions). The first line represents the irreducible decomposition \citep{Olive2017EffectiveCO} of the equivariant feature $h_i^l$. The Gate operation on spherical harmonic types of tensors is inherited from \cite{Thomas2018TensorFN, Gong_2023}. After obtaining the equivariant edge message $m_{ij}$, the subsequent steps mirror the vertex update module in \cite{Gong_2023}.

Observing the formulas, we note that the node features $h_i^l$ are equivariant spherical harmonics. The edge update then performs the tensor product between $h_i^l$, $h_j^l$, and the edge spherical harmonic basis $Y(\mathbf{r}_{ij})$:
$$h_i^l\otimes h_j^l \otimes Y(\mathbf{r}_{ij}).$$
By the CG-decomposition, the output of the tensor product can be decomposed back into spherical harmonics, aligning with the scalarization step by vector frames. Unlike Clofnet and Painn, \Model{}\_DeepH-E3 achieves higher-order tensorization by projecting scalars into higher-order spherical harmonic bases.

\end{document}